\documentclass[letterpaper, 10 pt, conference]{ieeeconf}  

\IEEEoverridecommandlockouts                              

\usepackage{cite}
\usepackage{amsmath,amssymb,amsfonts}
\usepackage{algorithmic}
\usepackage{graphicx}
\usepackage{arydshln}
\usepackage{textcomp}
\usepackage{xcolor}
\usepackage{booktabs}
\def\BibTeX{{\rm B\kern-.05em{\sc i\kern-.025em b}\kern-.08em
    T\kern-.1667em\lower.7ex\hbox{E}\kern-.125emX}}

\title{\LARGE \bf
Enhanced Anomaly Detection for Capsule Endoscopy Using Ensemble Learning Strategies*
}

\author{Julia Werner$^{1}$, Christoph Gerum$^{1}$, Jörg Nick$^{2}$, Maxime Le Floch$^{3,4}$,\\ Franz Brinkmann$^{3,4}$, Jochen Hampe$^{3,4}$, and Oliver Bringmann$^{1}$
\thanks{*This work has been partly funded by the German Federal
Ministry of Education and Research (BMBF) in the project MEDGE (16ME0530).}
\thanks{$^{1}$ Department of Computer Science, University of Tübingen, Tübingen, Germany}
\thanks{$^{2}$ Seminar of Applied Mathematics, ETH Zürich, Zürich, Switzerland}
\thanks{$^{3}$ Else Kröner Fresenius Center for Digital Health, TU Dresden, Dresden, Germany}%
\thanks{$^{4}$ Department of Medicine I, University Hospital Dresden, TU Dresden, Dresden, Germany}
}

\begin{document}

\maketitle
\thispagestyle{empty}
\pagestyle{empty}
\begin{abstract}

Capsule endoscopy is a method to capture images of the gastrointestinal tract and screen for diseases which might remain hidden if investigated with standard endoscopes. Due to the limited size of a video capsule, embedding AI models directly into the capsule demands careful consideration of the model size and thus complicates anomaly detection in this field. 
Furthermore, the scarcity of available data in this domain poses an ongoing challenge to achieving effective anomaly detection.

Thus, this work introduces an ensemble strategy to address this challenge in anomaly detection tasks in video capsule endoscopies, requiring only a small number of individual neural networks during both the training and inference phases. Ensemble learning combines the predictions of multiple independently trained neural networks. This has shown to be highly effective in enhancing both the accuracy and robustness of machine learning models. However, this comes at the cost of higher memory usage and increased computational effort, which quickly becomes prohibitive in many real-world applications. Instead of applying the same training algorithm to each individual network, we propose using various loss functions, drawn from the anomaly detection field, to train each network. The methods are validated on the two largest publicly available datasets for video capsule endoscopy images, the Galar and the Kvasir-Capsule dataset. We achieve an AUC score of 76.86\% on the Kvasir-Capsule and an AUC score of 76.98\% on the Galar dataset. Our approach outperforms current baselines with significantly fewer parameters across all models, which is a crucial step towards incorporating artificial intelligence into capsule endoscopies.
\end{abstract}

\section{Introduction}

Gastrointestinal (GI) diseases can affect the life quality severely. While some are easier to detect, others remain hidden until the consequences are too severe. Regions such as the mouth, esophagus, stomach as well as the colon can be successfully evaluated by standard techniques, such as a gastroscopy or colonoscopy~\cite{schindler1937gastroscopy, williams1973colonoscopy}. However, the small intestine is predominantly inaccessible by such endoscopes. A minimal invasive method that has been emerged since the early 2000s is the capsule endoscopy, which can visualize this part of the GI tract~\cite{iddan2000wireless, enns2017clinical, costamagna2002prospective}. With a video capsule endoscopy (VCE), a patient ingests a small pill-sized capsule comprising of an integrated camera and a light-emitting diode. As it traverses the GI tract through peristalsis, the capsule transmits recorded images to an electronic device~\cite{mylonaki2003wireless}. Afterwards, this enables the identification of anomalies within the duodenum, jejunum and the ileum, which correspond to the distal first section, the midsection and the last section of the small intestine, respectively~\cite{johnson2006physiology}. However, given the limited size of such video capsule, incorporating artificial intelligence (AI) directly into the capsule requires careful consideration of the overall model size and thus complicates anomaly detection on-site. 
Additionally, due to the limited data availability in this field, successful anomaly detection remains challenging in general. 
A classical approach to improve the performance of machine learning methods is ensemble learning~\cite{sagi2018ensemble, vanerio2017ensemble,dong2020survey}. It integrates multiple models and hereby balances the weaknesses of individual models. 

To leverage this, in this work, ensemble learning is applied to VCE data by combining different anomaly detection methods while considering the overall model size compared to the current state-of-the-art methods in this field. This provides a starting point to construct hardware-aware models enabling on-site anomaly detection. Inference on the hardware of the capsule endoscopy is particularly desirable for adaptive sampling rates and transmission. Upon anomaly detection, only the picture of interest should be transmitted for further examination by specialized physicians. Through this on-site evaluation, the required energy of such a capsule can be drastically reduced since the costly transmission of unproblematic images is omitted. Furthermore, this would allow the determination of anomalies in real-time and for example, to act immediately upon detection by increasing the resolution or the frame rate on-site at the region of interest. As a first step, this work transfers well-established anomaly detection techniques to this specific medical setting and explores reliable practices to combine those. Within this process, the total number of parameters of the proposed models as well as the feasibility to implement this in hardware, is kept in mind. 

\textbf{Our contribution.} In this paper, we describe a methodology for the construction of hardware-efficient ensemble learning models for anomaly detection tasks. We achieve this goal by formulating different anomaly detection algorithms based on the hardware-efficient MobileNet architecture, which are then combined in a simple ensemble method procedure to construct efficient anomaly detection algorithms. The proposed methodology produces models that outperform state-of-the-art models on the two largest video capsule endoscopy datasets, while requiring only a fraction of the storage memory cost compared to the state-of-the-art methods in this medical field. With these results, we aim to provide a key contribution towards the on-edge application of machine learning models for capsule endoscopies.

\section{Background and Related Work}
In this section, the main components of the present work are described, namely background on ensemble learning, anomaly detection methods as well as the current state of publicly available datasets comprising of video capsule endoscopy studies.

\subsection{Ensemble Learning}
Ensemble learning can be employed to combine various machine learning methods and thereby, to better capture the underlying structure of data~\cite{sagi2018ensemble, vanerio2017ensemble, dong2020survey}. It has been used for anomaly detection in a number of fields, for example in network security~\cite{vanerio2017ensemble, shahzad2022cloud} or in medicine  for neurocognitive disorder detection based on MRI datasets~\cite{savio2011neurocognitive}. 

Models based on ensemble learning often provide superior performance compared to their individual building blocks - at the cost of additional parameters and significant more operations during inference. Following \cite{H15}, different techniques have been presented to accelerate the inference of ensemble models, see e.g. \cite{Wen20}. Here, we aim to alter the training methods on the same architecture and construct methods that require only very few neural network evaluations during inference. We therefore do not require additional model order reduction techniques at this stage, although these ideas might provide avenues for future research.

\subsection{Anomaly Detection}
Anomaly detection remains a critical challenge in a wide range of applications, such as cyber security, network intrusion detection or health care~\cite{chandola2009anomaly}. Many methodologies evolved across various fields of machine learning that have been proven effective for outlier detection. In the field of supervised machine learning, models are trained on labeled data only. However, this relies on the availability of preferably large labeled datasets and especially in the field of medicine, this is less suitable due to the scarcity of labeled data. Unsupervised techniques form an alternative by exploiting the existence of unlabeled data, e.g. by employing deep autoencoders~\cite{zong2018deep}. Semi-supervised techniques additionally make use of unlabeled as well as normal data~\cite{akcay2019ganomaly}. In this work, we aim to harness the strengths of each area (supervised, unsupervised, semi-supervised) to maximize the use of the limited available data, while keeping in mind that only small networks can ultimately be deployed on hardware for this application.\\

\subsection{Hardware-Aware Machine Learning Methods}
The implementation of machine learning methods on edge devices remains a challenge, in particular when strong energy constraints are imposed on the available hardware. The development of hardware aware machine learning methods is therefore an active field of research and continuously evolving. 
A successfull family of network architectures based on depthwise separable convolutions is the MobileNet architecture ~\cite{howard2017mobilenets}, which was explicitly designed for embedded and mobile devices. In many real-world applications, these networks have demonstrated competitiveness with much larger models while using significantly fewer parameters.
Finally, low-power hardware accelerators, explicitly designed for the efficient inference of neural network architectures, have been shown to significantly reduce power usage and inference time for restricted neural network architectures \cite{bernardo2020ultratrail}. 

In this paper, we combine the progress made in the development of these architectures with anomaly detection and ensemble learning techniques, to construct hardware aware methods for video capsule endoscopies. In particular, for the medical investigation of the gastrointestinal tract only methods suitable for real-time low-power hardware architectures can be considered when neural networks are transferred to the real-world application.

\subsection{Datasets for Capsule Endoscopy}
The presented methods have been developed for the following two datasets, which are the largest publicly available video capsule datasets. In general, video capsule endoscopy datasets are sparse but remain essential if one wants to target pathological frames within capsule endoscopies.

\begin{figure*}[htpb]
\centering
\begin{minipage}{0.98\textwidth}
\includegraphics[width=0.49\linewidth]{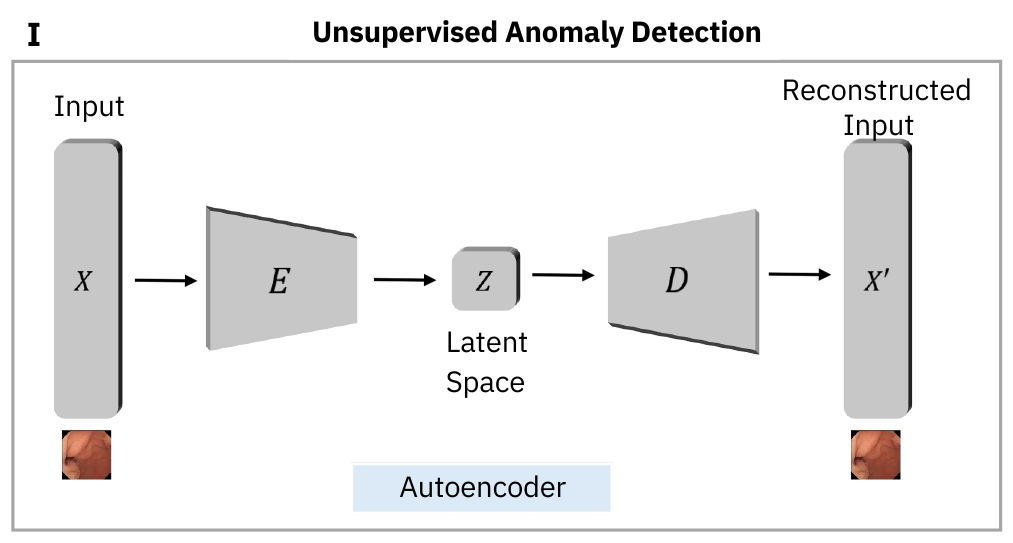}
\includegraphics[width=0.49\linewidth]{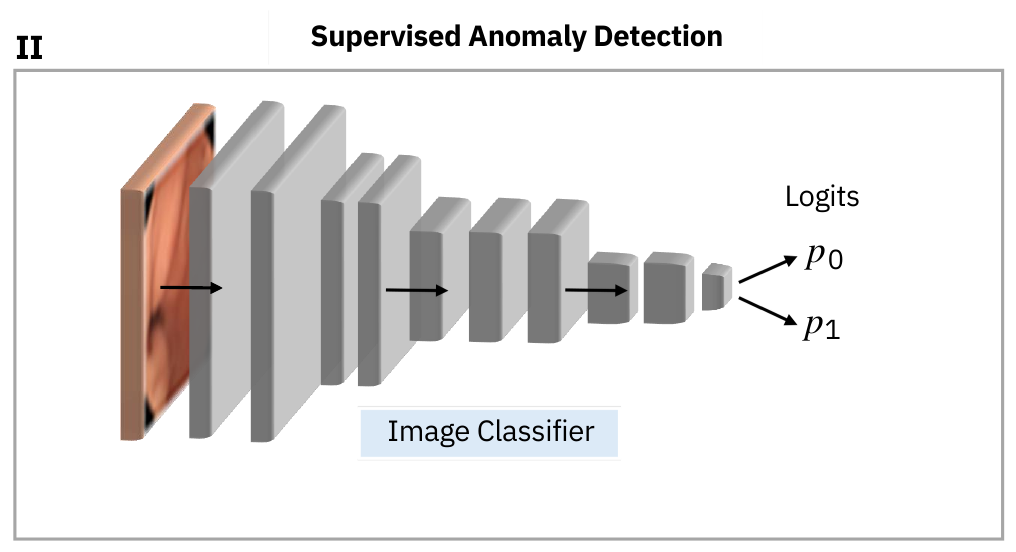}\\
\includegraphics[width=0.49\linewidth]{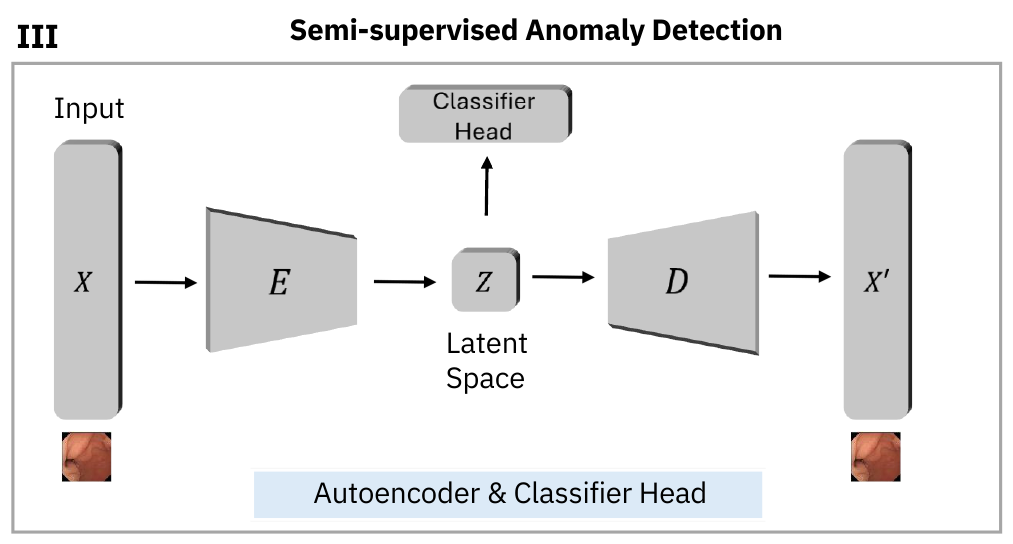}
\includegraphics[width=0.49\linewidth]{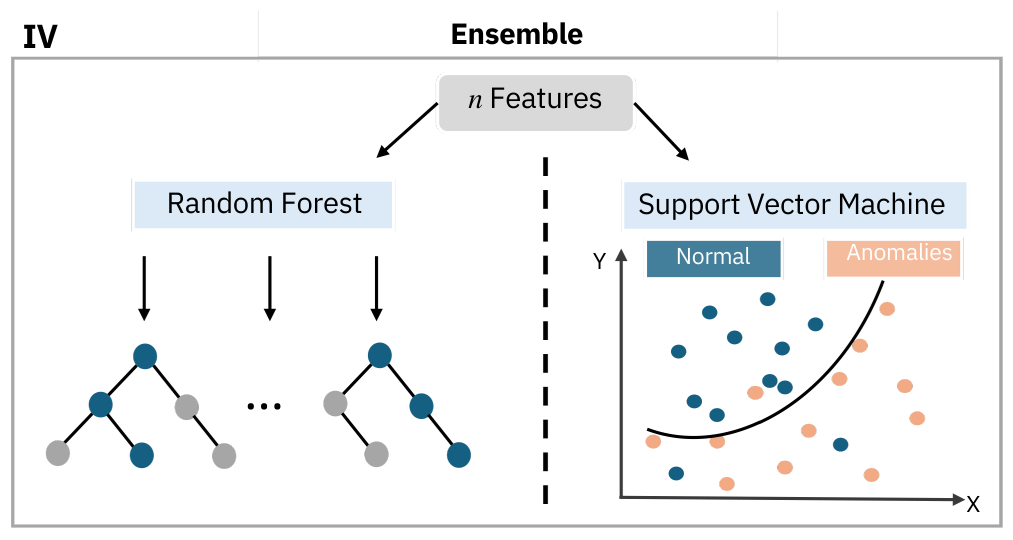}
\end{minipage}
    \caption{Overview of the ensemble learning strategy for Anomaly detection, consisting of an unsupervised (I), a supervised (II) and a semi-supervised (III) classification approach, followed by an ensemble model constituting of either a random forest model or a SVM (IV). I, II and III are all based on the same network architecture.}
    \label{fig:overview_ensemble}
\end{figure*}

\textbf{Kvasir-Capsule.} In 2021, the video capsule endoscopy dataset Kvasir-Capsule Dataset~\cite{smedsrud2021kvasir}, has been published. To our knowledge, representing the only publicly available large video capsule dataset for several years with a total of $47,238$ labeled and $4,694,266$ unlabeled frames. As mentioned by the authors, it is crucial to use the official split as published~\cite{smedsrud2021kvasir}. Simply splitting all available images does not ensure that images from the same patient appear only in the training but not in the test set. Due to the close similarity of frames, one would need to perform splits on the patient studies. Thus, we only compare results from this work with research that stuck to the official given splits or specifically state that they performed a split patient-wise.

\textbf{Galar Dataset.}
Recently, in 2024, a new large multi-label video capsule endoscopy dataset was published: Galar~\cite{le2025galar}, which provides a total of $3,513,539$ labeled images from $80$ patient VCE studies in total. This dataset contains annotated images with a variety of classes, including anatomical as well as pathological findings. 
    
In the following, we propose a combination of methods leading to improved results for both datasets while using notably smaller models. Employing smaller networks is crucial if one aims for execution on low-power hardware architectures as employed for capsule endoscopy, which is predominantly limited by its capsule size. In this work, we validate and compare our methods on those two largest publicly available VCE datasets.

\section{Methodology and Experiments}
Medical datasets are expensive to label and therefore often contain large amounts of unlabeled data. Nevertheless, significant progress has been made in the acquisition of labeled medical data, as is demonstrated by the two capsule endoscopy datasets treated in this paper.

Our method therefore makes use of the substantial amount of labeled data present as well as the vast amount of unlabeled data available. Consequently, we combine three different types of classifiers: Supervised anomaly detection, Semi-supervised anomaly detection and unsupervised anomaly detection.

Figure~\ref{fig:overview_ensemble} illustrates the overall classification approach. It comprises four parts, (I) an autoencoder, representing the unsupervised technique, (II) a standard image classifier drawn from the supervised methods and (III) an autoencoder in combination with an additional classification head, which forms the semi-supervised approach. The final ensemble model (IV), consisting of either a Random Forest model or a Support Vector Machine (SVM), combines the prediction of each method and returns the final evaluation.

\subsection{Datasets and Preprocessing}

The main goal of this proposed pipeline is the anomaly detection of VCE images, especially in the region of interest, the small intestine. To achieve this, the classes of the Kvasir-Capsule dataset were categorized into two classes, normal and anomaly, for binary classification: 
\begin{itemize}
    \item \textbf{Normal:} Pylorus, Reduced Mucosal View, Ileo-cecal valve, Normal Clean Mucosa
    \item \textbf{Anomaly}: Angiectasia, Blood-fresh, Foreign Bodies, Ulcer, Erosion, Lymphangiectasia.
\end{itemize}

\begin{figure}
  \begin{center}
\includegraphics[scale=0.45]{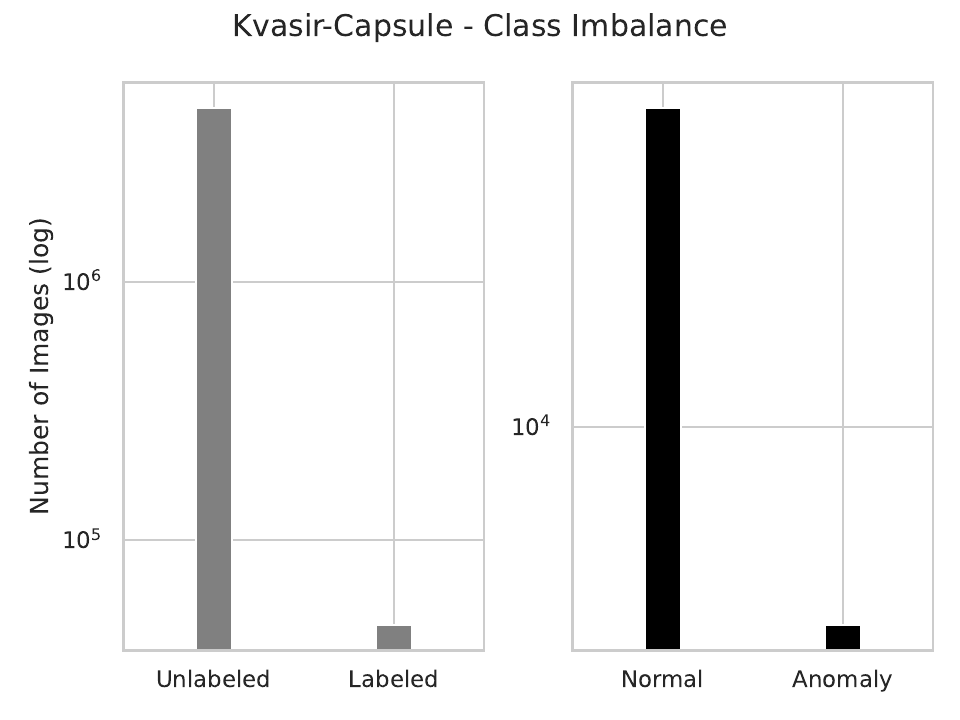}
  \end{center}
  \caption{\label{fig:kvasir_freq} Visualization of the number of images available in the Kvasir-Capsule dataset (unlabeled vs. labeled and normal vs. anomaly).}
\end{figure}

Figure~\ref{fig:kvasir_freq} visualizes the huge class imbalances within the Kvasir-Capsule dataset. While there are approximately $4,7$~M unlabeled frames, within the $\approx 47,000$ labeled frames, only about $9 \%$ are pathological images. This highlights the importance of employing machine learning methods that specifically target such outliers.

The Galar dataset as originally published is partitioned into different splits for each individual pathology but not for multiple pathological classes jointly. In order to perform anomaly detection, new splits for the training and validation sets were generated patient-wise, assembling the corresponding following pathology classes drawn from the small intestine only: 
\begin{itemize}
    \item \textbf{Normal:} Normal Clean Mucosa
    \item \textbf{Anomaly}: Polyp, Blood, Active Bleeding, Angiectasia, Erosion, Erythema, Ulcer.
\end{itemize}
The test set still comprises the same patient studies as originally published (patient IDs $61-80$). The remaining patient studies (ID $1-60$) were split in a $80:20$ (train : val) ratio.

Depending on the camera, the videos were originally recorded with a varying resolution ranging from $256 \times 256$ to $512 \times 512$~\cite{smedsrud2021kvasir} for Kvasir-Capsule and from $336 \times 336$ to $576 \times 576$ pixels for the Galar dataset~\cite{le2025galar}, which we initially adapted. However, since reducing the image sizes can lead to a model with lower complexity, we reduced the resolution for training the images to $224 \times 224$ as also employed by~\cite{le2025galar}, while keeping the original image sizes for testing. We found that reducing the image size to $224 \times 224$ does not impair the overall model accuracy. For data augmentation, random rotation, random vertical and horizontal flip as well as random erasing of small parts of images was applied.

\subsection{Autoencoder - Unsupervised and Semi-Supervised Anomaly Detection}
As demonstrated in the past, autoencoders do not only have remarkable reconstruction capabilities but are furthermore effective for outlier detection~\cite{an2015variational, zhou2017anomaly, gong2019memorizing}. They are also beneficial in a semi-supervised setting, by enabling the incorporation of unlabeled data. Since the given Kvasir-Capsule dataset has millions of unlabeled data, we employ autoencoders to leverage this.
    
An autoencoder $A$ is trained such that the decoder $D$ learns to reconstruct a given input $X$ which is encoded by an encoder~$E$. Thus, the following optimization problem needs to be solved
\begin{align}
        \min_{D,E} = \lVert X-D(E(X))\rVert,
\end{align}
with $\lVert \cdot \rVert$ as the $l2$-norm~\cite{zhou2017anomaly}.

The mean squared error (MSE)~\cite{hecht1992theory} was employed as a reconstruction loss in all autoencoder experiments:
\begin{align}
    \mathrm{MSE} = \frac{1}{N} \sum_{i=1}^N (y_i -\hat{y}_i)^2,
\end{align}
and is directly used as a feature in the ensemble model.
One special characteristic in this setting is that the autoencoder is based on the same network architecture as has been used for the image classifier. More precisely, the encoder consists of the MobilNetV3-Small architecture. According to this, the decoder was generated with the PyTorch framework
and the pretrained model fine-tuned for additional $15$ epochs. Data augmentation is employed to reduce overfitting. Hence, by training the autoencoder only with unlabeled or normal samples, we tried to emphasize a good reconstruction capability of normal samples and an inferior reconstruction performance for unseen anomalies. In the semi-supervised setting, for training, unlabeled as well as labeled images were used for the Kvasir-Capsule dataset. The Galar dataset only comprises labeled frames, thus, only labeled data was incorporated. Additionally, the compressed data from the latent space was processed with a small classifier head and its final prediction also used as a feature for the ensemble model.

\subsection{Image Classification - Supervised Anomaly Detection}

MobileNets have been successfully applied to image classification and recognition tasks, with the added advantage of being highly efficient on mobile as well as embedded devices~\cite{howard2017mobilenets, howard2019searching, chiu2020mobilenet}. This makes them an ideal solution for the given problem, as they are well-suited for low-power hardware while still delivering strong classification performance. In the past, others have generated FPGA-based accelerators specifically designed for this architecture and proven its efficiency~\cite{yan2021fpga, liao2019design}.
Therefore, as the most light-weight approach, standard image classification was performed by fine-tuning a pretrained MobileNetV3 for $15$ epochs with the Adam optimizer, 
using PyTorch~\cite{Ansel_PyTorch_2_Faster_2024} and the HANNAH framework~\cite{gerum2022hardware}. Additionally, a weighted sampler was employed to address the omnipresent class imbalances of these two datasets.
In this supervised setting, training was only performed with labeled anomaly and normal VCE images by employing the cross-entropy (CE)~\cite{zhang2018generalized,Ansel_PyTorch_2_Faster_2024} loss
\begin{align}
    \mathrm{CE} = -\frac{1}{N} \sum_{i=1}^N y_i \log(p(y_i)) + (1-y_i)\log(1-p(y_i)).
\end{align}

\subsection{Ensemble Model}
The key concept of the presented methodology is to provide an ensemble method that yields an overall improved prediction performance and therefore outperforms the individual methods by combining all given models.
As described by~\cite{sagi2018ensemble}, for a dataset with $m$ features and $n$ samples $D = \{(x_i, y_i)  \, \mid \, x_i \in \mathbb R^m, y_i, \in \{0,1\}, \, i=1,\dots,n\}$, the classification of an ensemble model $\phi$ with $K$ base learners $\{f_0, f_1, \dots,f_k\}$ can be described by the following function:
\begin{align*}
            \hat{y}_i = \phi(x_i) = H(f_0, f_1,...,f_k),
\end{align*}
with $\hat{y}_i \in \mathbb Z$ for classification problems. For our setting, we apply $k=3$ different base learners (autoencoder, autoencoder with classification head and image classifier) and two different ensemble models~$\phi$ (Random Forest Classifier~\cite{breiman2001random} or a SVM~\cite{chen2005tutorial, hearst1998support}, with $\hat{y}_i \in \{0, 1\}$.
A random subset of the training and validation sets were used to perform a random search and find the best parameters for each ensemble model and to finally train those models. For final evaluation, the test set was used.

\section{Results}

The results of the experiments performed on the Kvasir-Capsule dataset are listed in Table~\ref{tab:results_kvasir}. 
For the 12-class problems treated in recent research, the macro-average, the micro-average and weighted metrics are shown, if accessible. If someone aims to find all pathological images, the macro-average weighted metrics are considered to be the most relevant ones for the multi-class settings with large class imbalances as the different classes are weighted equally. For the micro-average and the weighted metrics, the larger class is given more weight, which is not of interest in anomaly detection since the larger class typically consists of the normal samples. However, for the purpose of completeness, both metrics are shown, if accessible. For binary-class problems, such distinction is not relevant, which impedes the comparison across these studies. Thus, for the final comparison, we consider only the results presented below the dashed line. Nevertheless, those baseline results are also shown in order to provide an overview of essential results in this field. 

\begin{table}[htbp]
	\caption{Results of the ensemble model on the Kvasir-Capsule dataset compared to the baseline results.}
            \begin{center}\resizebox{\columnwidth}{!}{
			\begin{tabular}{ccccccc}
					\toprule
                    \textbf{Method} & \textbf{AUC} &  \textbf{Recall} & \textbf{Accuracy}  & \textbf{F1 Score} & \textbf{MCC} &  \textbf{Precision}\\
                    \midrule\midrule
                    \textbf{Baseline Results}\\
                    \midrule\midrule
                    DenseNet161, Baseline~\cite{smedsrud2021kvasir} - macro & -- & 28.12 & -- & 25.60 & 43.29 & 29.94\\
                    DenseNet161, Baseline~\cite{smedsrud2021kvasir} - micro & -- & 73.66 & -- & 73.66 & 43.29 & 73.66\\
                    FocalConvNet~\cite{srivastava2022video}  - macro  & -- & 27.45 & 63.73 & 21.78 & 29.64 & 24.38\\ 
                    FocalConvNet~\cite{srivastava2022video} - weighted & -- & 63.73 & 63.73 & 67.34 & 29.64 & 75.57\\ 
                    ViT~\cite{regmi2023vision} - weighted & 57.0 & 71.56 & 71.56 & 71.56 & 37.05 & 68.41\\[1mm] \hdashline \\
                    ResNet152 and OneClassSVM~\cite{de2023abnormality} & - & 56.00 & -& 50.00 & - & 55.00\\
                    ResNet152 and XGBoost~\cite{de2023abnormality} & - & 56.00 & -& 57.00 & - & 73.00\\
                    \midrule\midrule 
                    \textbf{This Work}\\
                    \midrule\midrule
  Image Classifier (CLF) & 73.42 &  51.96 & 91.65 & 48.30 &  43.91 & 45.12\\ 
                    Autoencoder (AE) & 64.70 & 40.90 & 84.94 & 28.96 &  22.54 &  22.42 \\  
                    Ensemble SVM, AE, CLF & 76.80 & 61.59& 89.72 & 47.36 & 43.43 & 38.47\\
                    Ensemble RF, AE, CLF & 76.86& 60.65 & 90.64& 49.32 & 45.33 & 41.56\\ 
					\bottomrule
				\end{tabular}}
\end{center}
\label{tab:results_kvasir}
\end{table}

The authors from the Kvasir-Capsule dataset~\cite{smedsrud2021kvasir} yield a macro-average precision of $29.94\%$, a recall of $28.12\%$ and a F1-score of $25.60\%$ with a DensNet-161. Srivastava et al.~\cite{srivastava2022video} achieved an accuracy of $63.73\%$ and a macro-average recall of $27.45\%$ with a FocalConvNet including depth-wise separable convolutions followed by a GeLU activation layer. 
From those authors listed above, only de Sá et al~\cite{de2023abnormality} also performed a binary study, splitting the Kvasir-Capsule dataset into binary classes in a similar way as conducted in this work and is therefore ideally suited for comparison. 
They trained a ResNet152 and fine-tuned it with XGBoost and an OneClassSVM. They report a recall of $56\%$, a F1-score of $57\%$ and a precision of $73\%$ while requiring about $\approx 60$ million parameters.

Table~\ref{tab:results_kvasir} demonstrates that both ensemble models outperform the individual methods in terms of the AUC score ($76.86\%$), sensitivity ($60.65\%$)
and Matthew's correlation coefficient (MCC) ($45.33\%$). The accuracy as well as the precision is higher if only the image classifier is used. However, considering the low sensitivity, this is most likely due to a large fraction of true negatives while not detecting many true positives. Overall, both ensemble models involving either the SVM or Random Forest perform well compared to \cite{de2023abnormality} by detecting a larger fraction of true positives ($\approx 61\%$ vs. $56\%$). This partially probably comes at the cost of a lower precision compared to \cite{de2023abnormality}. Importantly, while demonstrating a good performance, the total model sizes of all presented models is drastically reduced from around $60$ million to a maximum of $4$ million parameters.

\begin{table}[htbp]
	\caption{Results of the ensemble model on the Galar dataset.}
            \begin{center}\resizebox{\columnwidth}{!}{
			\begin{tabular}{ccccccc}
					\toprule
                    \textbf{Method} & \textbf{AUC} &  \textbf{Recall} & \textbf{Accuracy}  & \textbf{F1 Score} & \textbf{MCC} &  \textbf{Precision}\\
                    \midrule\midrule
                    Image Classifier (CLF) & 74.94 &  60.88& 87.28 & 37.01 &  34.45 & 26.59\\ 
                    Autoencoder (AE) & 45.83 & 28.94& 60.65 & 8.28 &  -4.15 &  4.83 \\  
                    Ensemble SVM, AE, CLF & 76.98 & 80.58 & 73.83& 27.44& 28.29 &16.53\\
                    Ensemble RF, AE, CLF & 68.34& 86.06 & 52.80 & 18.3 & 17.62 &  10.24\\ 
					\bottomrule
				\end{tabular}}
\end{center}
\label{tab:results_galar}
\end{table}

Following this, the proposed approach was validated on anomalies in the small intestine from the Galar dataset. Since this dataset has just been recently published, there are no results published targeting the pathologies jointly. 
The image classifier yields the highest scores in terms of accuracy, F1 score, MCC score and precision. From our proposed models, the ensemble model including the SVM attains the highest AUC score ($76.98\%$). The ensemble model including the random forest yields the highest recall ($86.06\%$), but has an inferior accuracy compared to the image classifier. Interestingly, in contrast to the Kvasir-Capsule dataset, the Random Forest ensemble model is inferior compared to the SVM ensemble model on the Galar dataset. Overall, the image classifier shows a strong performance while the autoencoder exhibits inferior results. One possible reason, in comparison to the Kvasir-Capsule dataset, is that the availability of a substantial amount of labeled data allows the image classifier to be well-trained independently. However,  without any unlabeled data, the strengths of the autoencoder can not be fully exploited. Building on the successful classification of individual anomaly tasks in \cite{le2025galar}, we present the first results on anomaly detection with the majority anomaly classes pooled together from this dataset, offering a potential foundation for future work.

\begin{figure}[htbp]
		\centering\includegraphics[width=0.8\linewidth]{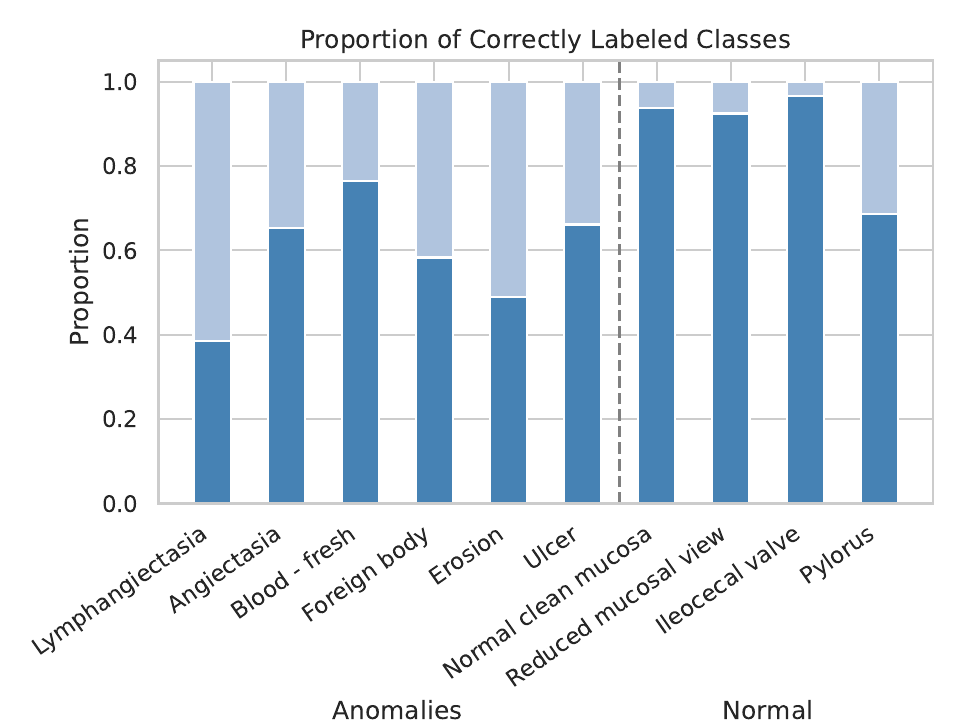}
			\caption{Proportion of correctly labeled classes of the Kvasir-Capsule dataset if evaluated with the ensemble model including the Random Forest classifier, the autoencoder and the image classifier.
				\label{fig:proportion_kvasir}}
\end{figure}

\begin{figure}[htbp]
		\centering\includegraphics[width=0.8\linewidth]{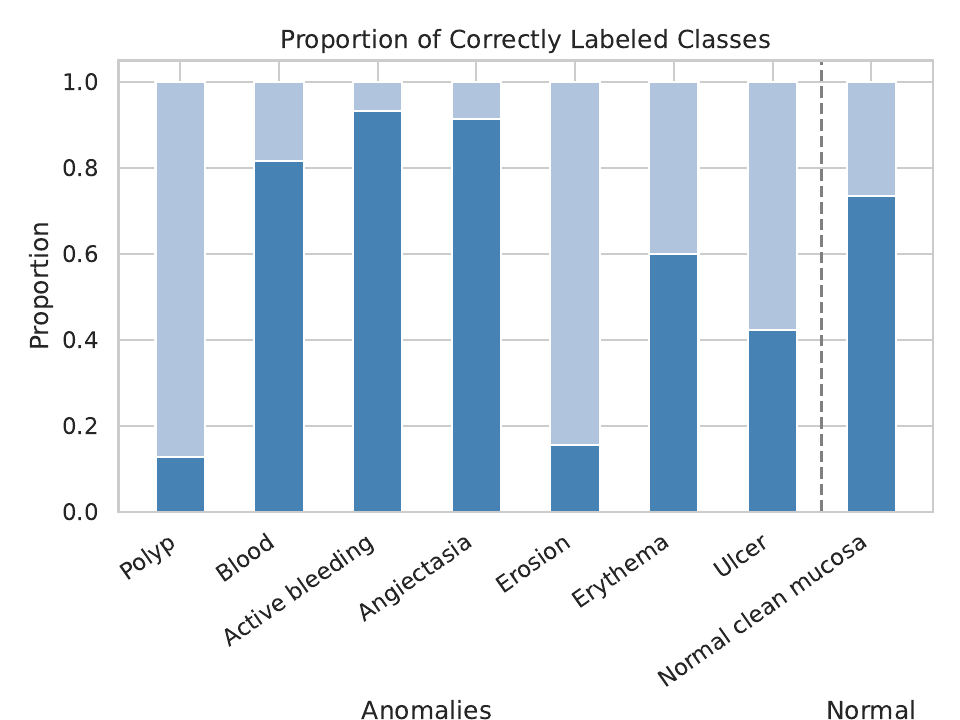}
			\caption{Proportion of correctly labeled classes of the Galar dataset if evaluated with the ensemble model including a SVM, the autoencoder and the image classifier .
				\label{fig:proportion_galar}}
\end{figure}

Additionally, we explored how well the individual pathologies were detected by plotting the proportion of correctly labeled classes for both datasets for one of the ensemble models exemplarily. For the Kvasir-Capsule dataset (Figure~\ref{fig:proportion_kvasir}), among the normal samples, the pylorus class has the highest misclassification rate. Among the anomalies, the classes lymphangiectasia and erosion exhibited the highest error rates. The misclassification of the pylorus might be explained with the occasional occurence of anomalies such as erosions or redness around the pylorus region which have similarities with the small intestine pathologies and thus, might be recognized as such by the classifier. Figure~\ref{fig:proportion_galar} reveals huge differences in the proportion of correctly labeled classes for the galar dataset. While blood, active bleeding and angiectasias are surprisingly well detected, polyps and erosions are mostly not correctly classified. It’s possible that these well-detected classes share certain properties that the ensemble identifies as anomalous.


\begin{figure}[htbp]
		\centering\includegraphics[width=0.8\linewidth]{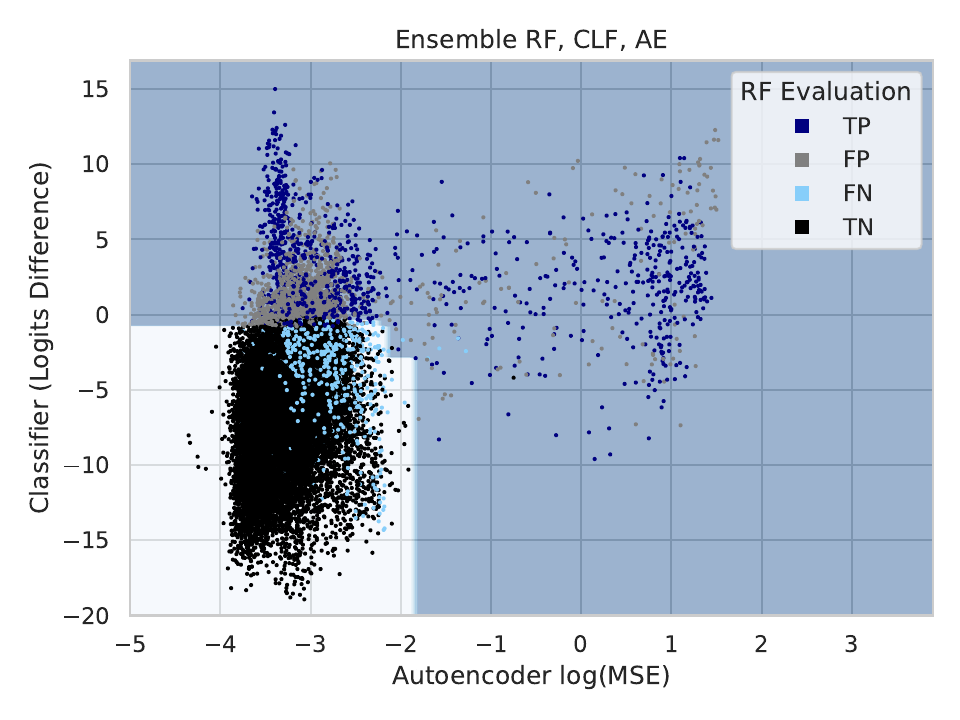}
			\caption{Classification results of the ensemble model including the Random Forest classifier. The logits distance of the image classifier is plotted over the $\log(\text{MSE})$ of the autoencoder and the true positives (TP), true negatives (TN), false positives (FP) and false negatives (FN) labeled by color as evaluated by the ensemble model. The blue area indicates the region in which the ensemble model classifies data samples as anomalous.
				\label{fig:scatter_rf}}
\end{figure}

The partial contribution of the individual components of the ensemble model and its overall advantage were particularly evident for the Kvasir-Capsule dataset. Thus, for this dataset, the classification results of the ensemble model including the random forest are more closely evaluated by plotting the distance of the logits of the image classifier over the $\log(\text{MSE})$ outputted by the autoencoder and finally marked as evaluated by the Random Forest model in Figure~\ref{fig:scatter_rf}. This shows, besides a large fraction of correctly classified true negatives, also a visible amount of false positives. However, detecting a large amount of true positives on the cost of some false positives is accepted in this setting because finding a majority of anomalies is prioritized. Furthermore, the shift of anomalies to the right side of the plot reveals that the autoencoder exhibits a higher loss on anomalies than on normal samples, as anticipated. We can further observe that the ensemble model recognizes that a larger MSE correlates with a higher possibility of actually being an anomaly.
\begin{figure}[htbp]
		\centering\includegraphics[width=0.75\linewidth]{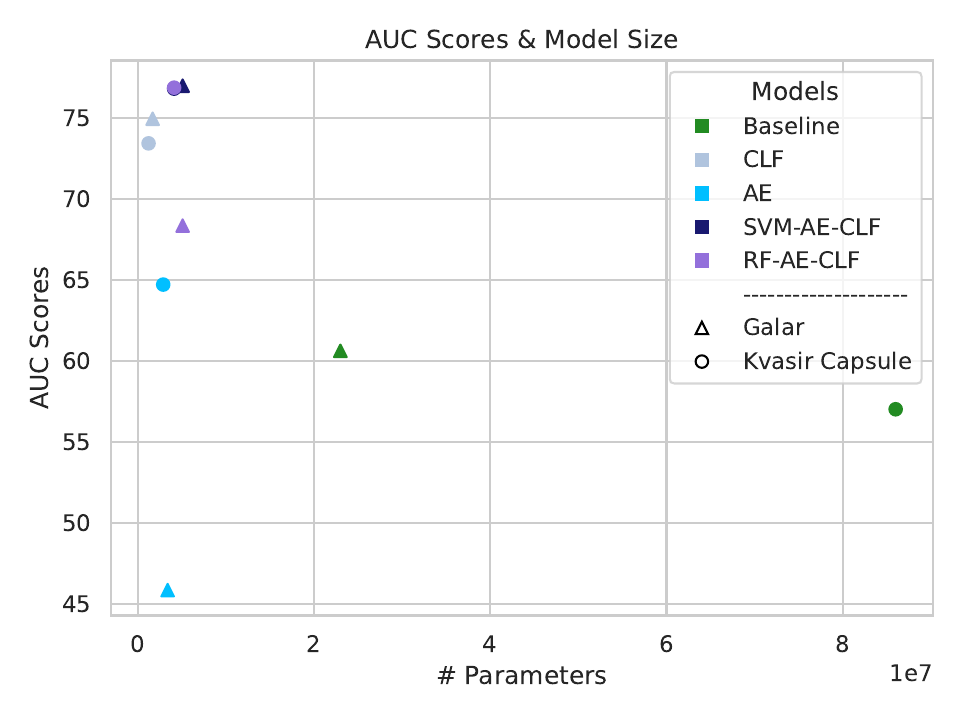}
			\caption{AUC scores shown in comparison to the number of parameters needed for each model evaluated using both datasets: Galar and Kvasir-Capsule (Please note that the SVM-AE-CLF marking for Kvasir-Capsule is overlayed by the RF-AE-CLF marking point).
				\label{fig:auc_plot}}
\end{figure}

Figure~\ref{fig:auc_plot} visualizes the classification performance of each model and for each dataset while comparing the number of parameters needed for each model. The ensemble model including a SVM demonstrates superior performance on the Galar dataset. For Kvasir-Capsule, the Random Forest ensemble model yields the best results. Importantly, all implemented methods need less parameters than the baselines while resulting in a higher AUC score.
More precisely, on the Kvasir-Capsule dataset we ultimately used around $4$ million parameters instead of the $86$ million parameters of the ViT from~\cite{regmi2023vision}. 
For the galar dataset, \cite{le2025galar} performed all experiments with a ResNet50 with around $25$ million~\cite{le2025galar} for single task pathology problems. We were able to reduce the number of parameters to $5$ million and additionally perform anomaly detection on multiple pathologies.

\section{Conclusion}
In this paper, we developed hardware-aware ensemble learning methods for anomaly detection and validated them on a critical real-world application: video capsule endoscopy (VCE). Previous studies have shown that anomaly detection with the two largest VCE datasets presents significant challenges. By using the same network architecture as the backbone for each component of the ensemble and constraining the total number of parameters, we produced significantly smaller models with enhanced classification performance compared to state-of-the-art models. This is a crucial step towards a suitable AI model capable of anomaly detection for capsule endoscopies.  Looking ahead, larger labeled datasets of pathological images from the small intestine are essential, as the current scarcity of data continues to impede progress in anomaly detection and remains a major bottleneck.

\bibliographystyle{splncs04}
\bibliography{references}

\end{document}